\documentclass[conference]{IEEEtran}
\IEEEoverridecommandlockouts

\usepackage[utf8]{inputenc}
\usepackage{amsmath}
\usepackage{amssymb}
\usepackage[flushleft]{threeparttable}
\usepackage[
backend=biber,
style=numeric,
citestyle=ieee]{biblatex}
\usepackage{csquotes}

\usepackage[english]{babel}

\usepackage{graphicx}
\usepackage{float}

\makeatletter
\newcommand{\linebreakand}{%
  \end{@IEEEauthorhalign}
  \hfill\mbox{}\par
  \mbox{}\hfill\begin{@IEEEauthorhalign}
}
\makeatother

\begin{document}

\title{Behavioral graph fraud detection in E-commerce}

\date{July 2022}

\graphicspath{ {./behavior_sankey.png/} }



\author{\IEEEauthorblockN{1\textsuperscript{st} Hang Yin}
    \IEEEauthorblockA{
        \textit{eBay China}\\
        Shanghai, China \\
        hangyin@ebay.com}
    \and
    \IEEEauthorblockN{2\textsuperscript{nd} Zitao Zhang}
    \IEEEauthorblockA{
        \textit{eBay China}\\
        Shanghai, China \\
        zitzhang@ebay.com}
    \and
    \IEEEauthorblockN{3\textsuperscript{rd} Zhurong Wang}
    \IEEEauthorblockA{
        \textit{eBay China}\\
        Shanghai, China \\
        zhurowang@ebay.com}
    \and
    \IEEEauthorblockN{4\textsuperscript{th} Yilmazcan Özyurt}
    \IEEEauthorblockA{
        \textit{ETH Zürich}\\
        Zürich, Switzerland\\
        yozyurt@ethz.com}
    \linebreakand 
    \IEEEauthorblockN{5\textsuperscript{th} Weiming Liang}
    \IEEEauthorblockA{
        \textit{eBay China}\\
        Shanghai, China \\
        weimliang@ebay.com}
    \and
    \IEEEauthorblockN{6\textsuperscript{th} Wenyu Dong}
    \IEEEauthorblockA{
        \textit{eBay China}\\
        Shanghai, China \\
        wenydong@ebay.com}
    \and
    \IEEEauthorblockN{7\textsuperscript{th} Yang Zhao}
    \IEEEauthorblockA{
        \textit{eBay China}\\
        Shanghai, China \\
        yzhao5@ebay.com}
    \and
    \IEEEauthorblockN{8\textsuperscript{th} Yinan Shan}
    \IEEEauthorblockA{
        \textit{eBay China}\\
        Shanghai, China \\
        yshan@ebay.com}
}

\maketitle

\begin{abstract}
    %
    In e-commerce industry, graph neural network (GNN) methods are the new trends for transaction risk modeling. The power of graph algorithms lie in the capability to catch transaction linking network information, which is very hard to be captured by other algorithms. However, in most existing approaches, transaction or user connections are defined by ``hard link" strategies on shared properties, such as same credit card, same device, same ip address, same shipping address, etc. 
    Those types of strategies will result in sparse linkages by entities with strong identification characteristics (ie. device) and over-linkages by entities that could be widely shared (ie. ip address), making it more difficult to learn useful information from graph. To address aforementioned problems, we present a novel behavioral biometric based method to establish transaction linkings based on user behavioral similarities, then train an unsupervised GNN to extract embedding features for downstream fraud prediction tasks. To our knowledge, this is the first time similarity based ``soft link" has been used in graph embedding applications. To speed up similarity calculation, we apply an in-house GPU based HDBSCAN clustering method to remove highly concentrated and isolated nodes before graph construction. Our experiments show that embedding features learned from similarity-based behavioral graph have 
    achieved significant performance 
    increase to the baseline fraud detection model in various business scenarios. 
    In new/guest buyer transaction scenario, this segment is a challenge for traditional method, we can make precision increase from 0.82 to 0.86 at the same recall of 0.27, which means we can decrease false positive rate using this method.
\end{abstract}


\section{Introduction}

Fraud detection is an indispensable, critical function to e-commerce business. Majority of fraud risk models today are built using traditional statistic based features such as velocity, consistency, profile, etc. However, in our fraud trend analysis, we found a large number of attacks occurred in forms of fraud rings with sophisticated linkages. Such collective signals are not covered by existing individual features. Graph Neural Network (GNN), a cutting edge technology from the combination of deep learning and graph algorithms, is often found to 
provide better performance than other algorithms, given its capability to capture the relationships among linked users, transactions apart from their individual information. In most previous applications
of GNN algorithms to detect fraud
\cite{liang2018stole}\cite{liu2018heterogeneous}, 
the nodes are connected by some ``hard link" (real entity) strategies such as same device, etc. While in our transaction detection domain, such methods have some inherent short-comings due to the nature of data. Firstly, entities with strong identification characteristics such as devices, tend to result in very sparse linkages between transactions, making it hard to propagate risk information across the graph. While entities that could be easily shared by a large population such as public ip addresses, freight forward addresses, will result in concentrated hubs in graph. Fraudsters often take advantages of such hubs to increase the ambiguity of differentiating good and bad users. In this work, we propose an innovative idea of creating transaction-to-transaction  ``soft links" based on users behavioral biometric similarities, which is another strongly identifiable information and hard to fake. The behavioral biometric features are extracted from user page view sequences through an attention-lstm model. On top of the similarity soft link graph, we leverage GNN unsupervised training to learn 
node embeddings for each transaction node which are used as additional features for the downstream fraud prediction model. To make the similarity calculation more efficient during graph setup, we utilize our in-house GPU based HDBSCAN to cluster all behavioral features before establishing the similarity graph. This technique makes the graph building procedure 10x faster. We evaluate the effectiveness of graph embeddings in different downstream tasks, experiment results show in some particular populations, our proposed method is able to achieve significant improvement to the baseline model performance.

Our contribution can be summarized as:
\\(i)	We proposed a behavior biometrics based method for graph construction, users with similar behavior biometrics are connected in the graph.
\\(ii)	To speed up transaction graph building and make the graph more robust, we cluster all user behavior embeddings first, then drop isolated nodes and concentrated hubs, which makes it much faster to generate the final similarity graph.
\\(iii)	Based on the similarity soft-link graph, we leverage unsupervised GNN to aggregate similar nodes features. These features have been proved quite effective in some specific fraud detection scenarios, especially for new and guest buyer transactions which lack user historical information.

\section{Related works}
\subsection{Behavior sequence embedding}
For the past few years, numerous start-of-the-art sequence embedding approaches have been proposed and applied in modern machine learning systems to address a variety of tasks such as fraud detection 
 \cite{DBLP:journals/corr/abs-1808-05329}\cite{DBLP:journals/corr/abs-1912-11760}\cite{DBLP:explainableDeepBehaviorSeqClustering},
item recommendation \cite{DBLP:journals/corr/abs-1905-06874}\cite{DBLP:journals/corr/FangZC16}. 

In general, these methods aim to learn a representation from user behavior sequences where each sequence is composed of a series of events representing user actions, ie, user clicks view item page at timestamp $t_1$, clicks checkout page at time $t_2$. Except for the user action itself, there might be other attributes associated to some certain actions, for example in this work\cite{DBLP:journals/corr/abs-1808-05329}, a Markov Transition Field and LSTM-CNN backbone are used to extract feature. Various types of features at the same timestamp such as IP, browser language, URL are encoded separately and concatenated together to finally form a behavioral sequence of heterogeneous feature vectors, then an LSTM-CNN architecture is adopted to learn the behavioral embedding. In time attention based fraud transaction detection framework \cite{DBLP:journals/corr/abs-1912-11760}, a unified framework has been proposed to firstly embed user sequential behavior patterns, then combine learned embedding with static user profiles to train a supervised classification model through an end-to-end pipeline. In our analysis of transaction fraud prediction, dwell time of the action is very crucial in differentiating good and fraudulent users. An example is that at view item page, good users usually spend more time to review the item details before checkout, while fraudsters won't waste too much time on that page. In addition, we want to generate a more general representation of user behavior which can be used in different downstream tasks. 
Considering all the facts above, we refer to this, a time-attention based unsupervised method for behavior sequence embedding generation \cite{DBLP:explainableDeepBehaviorSeqClustering}. In this work, they proposed an unsupervised behavior sequence embedding network structure, also involved attention mechanism. But in this work, they clustered all the embeddings to identify fraud. While in our use case we only refer to the embedding generation part, and use embeddings to construct graph.

\subsection{High dimension clustering}
Many algorithms \cite{mccallum2000efficient} are present for clustering data. Partition-based techniques such as K-means \cite{4262534}, 
K-medoids \cite{PARK20093336}
use a selected distance measure to update the clusters in an iterative way until an optimal partition is obtained. Hierarchical clustering techniques such as CURE \cite{guha1998cure}
and CHAMELEON \cite{karypis1999chameleon} 
group data in a hierarchical manner. The hierarchical clustering can be visualised using a dendrogram which is an inverted tree that describes the order in which data points are merged (agglomerative hierarchical clustering ) or broken up (divisive hierarchical clustering). Density-based clustering such as DBSCAN \cite{ester1996density}
works by identifying “dense” clusters of points, allowing it to learn clusters of arbitrary shape and identify outliers in the data. Most clustering techniques require specifying a minimal distance or minimal sample number for clustering. However, in real world cases, it's always difficult 
\cite{steinbach2004challenges}
to determine these hyper-parameters. Furthermore, most existing clustering methods involve calculating distances between samples in the dataset. In case of large datasets of high-dimension vectors, it will inevitably take a long time to run the clustering. As we know, distance-based clustering methods start to fail as dimensionality increases. In high-dimension spaces, distribution of distances becomes denser and concentrated so that it's almost impossible to differentiate samples by distance. In such cases, we need to apply effective dimension reduction techniques to boost the performance of subsequent clustering algorithms.
UMAP \cite{mcinnes2018umap} is such a practical scalable method that applies to real world data which can be used for visualisation similarly to t-SNE, but also for general non-linear dimension reduction. When using this method there are some assumptions about the data, for example, the Riemannian manifold is locally connected. Based on this assumptions it is possible to model the manifold with a fuzzy topological structure. The embedding is found by searching for a low dimensional projection of the data with the closest possible equivalent fuzzy topological structure.

\subsection{Graph neural network}
In recent years, several graph neural network architectures \cite{abu2018watch}\cite{zhou2020graph} have been proposed to generate graph node embeddings for various applications.
The basic idea behind is to distill high-dimensional information about a node's graph neighborhood into a dense vector embedding. Graph convolutional networks (GCNs) which train a distinct embedding vector for each node, should be applied with a transductive setting on fixed graphs. To generalize the GCN \cite{DBLP:GCN}  approaches, GraphSage \cite{NIPS:GraphSAGE} has been proposed as an inductive unsupervised learning framework on large graph, which is able to efficiently generate node embeddings for previously unseen data. Learned embeddings capture the structural properties of a node's neighborhood that includes both the node's local information in the graph and its global position. In the unsupervised learning task, a graph-based loss function \eqref{eq.1} is designed to encourage nearby nodes to have similar representations, while enforce disparate node representations highly distinct.
\begin{equation}
\small 
    J_{\mathcal{G}}(\mathrm{z}_u)=-\log (\sigma (\mathrm{z}_u^{\top} \mathrm{z}_v)) - Q \cdot \mathbb{E}_{v_n \sim P_n (v)} \log (\sigma(-\mathrm{z}_u^\top \mathrm{z}_{v_n})) 
    \label{eq.1}
\end{equation}
where $v$ is a vertex near $u$ by a random walk of fixed length, $P_{n}$ is the probability distribution of negative samples and $Q$ is the number of negative samples. Unlike DeepWalk \cite{DBLP:deepwalk}, the vertex representation vector here is generated by aggregating neighbor features of the vertex, rather than simply performing an embedding lookup operation. Another key part of GraphSage is neighborhood sampling. In consideration of computational efficiency, a fixed number of neighbors are sampled to aggregate the representations of neighborhood for each node. Sampling without replacement is used to draw different uniform samples at each iteration. In case where the sample size is larger than the node’s degree, sampling with replacement is used until k nodes are sampled. In order to aggregate the features from neighbors, there are several candidate aggregator functions such as MEAN aggregator, Pooling aggregator and LSTM aggregator. In this paper, we choose MEAN aggregator to generate transaction embedding.

\section{Approach}
\subsection{Overview}

\begin{figure}[b]
\centering
\includegraphics[width=0.48\textwidth]{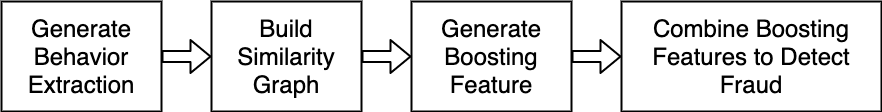}
\caption{Main steps of our proposed idea. First to get order's embedding and construct soft link graph, then get node embedding as added feature to the baseline model.}
\label{fig.pipeline}
\end{figure}

\begin{figure*}[t]
\centerline{\includegraphics[width=0.9\textwidth]{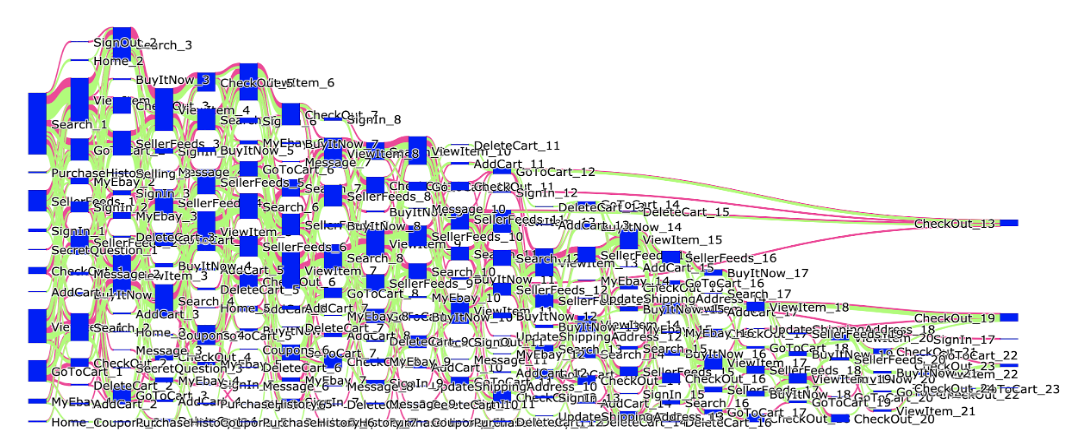}}
\caption{Page view behavior sequence}
\label{fig.behavior_sequence}
\end{figure*}

As talked above, constructing a transaction graph is a challenging task. Although a variety of properties such as shipping address, IP, email, and device, could be used to define the linking strategies, connections by different properties should be assigned with different weights to reflect their respective confidence levels, usually a lot of noise would be introduced by such method. On the other hand, if we only use strong linking properties such as devices and bank card tokens, connections in graph would be very sparse. Since user behavior sequence contains implicit identification information, we propose a new method that uses behavioral biometric similarity as ``soft links" between transactions. Based on constructed graph, we apply GNN algorithm to learn transaction node embeddings, which we called boosting features in the paper. The boosting features could be applied to downstream tasks of different risk management cases to further improve the performance of existing models. As the whole experiment process is complicated, we illustrate our proposed idea into following simplified flow chart in 
Fig. \ref{fig.pipeline} for better understanding. The input for Behavior extraction is user's transaction behavior sequence, and input for feature boosting part includes 100 profile and transaction related features.

\begin{figure}[t]
\centering
\includegraphics[width=0.48\textwidth,height=0.3\textwidth,keepaspectratio=FALSE]{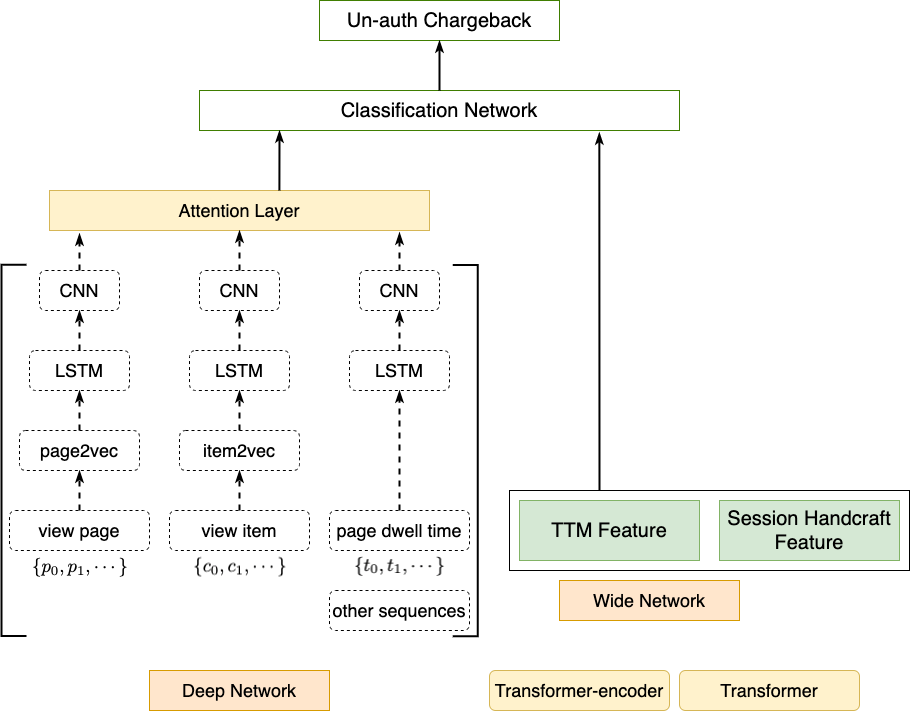}
\caption{Supervised network structure}
\label{fig.supervised_network_structure}
\end{figure}

\subsection{Generate user behavioral embedding}
According to our analysis for user page view sequence data on e-commerce website, we visualize different user behavioral patterns in Fig. \ref{fig.behavior_sequence}.
The red lines represent page view paths from fraudulent users, while the blue ones are for good users. We can see some certain transfer movements among fraudulent user paths. Therefore we try a supervised sequence embedding method at first. Given a user behavior sequence \{\textbf{\textit{s}}\textsubscript{\textit{i}}\}, where \textbf{\textit{s}}\textsubscript{\textit{i}} denotes user action at time step \textit{i} which consists of user features such as page ID \textbf{\textit{p}}\textsubscript{\textit{i}}, item category \textbf{\textit{c}}\textsubscript{\textit{i}}, page time \textbf{\textit{t}}\textsubscript{\textit{i}}, the sequences \{\textbf{\textit{p}}\textsubscript{\textit{i}}\}, \{\textbf{\textit{c}}\textsubscript{\textit{i}}\}, \{\textbf{\textit{t}}\textsubscript{\textit{i}}\} are sent to LSTM, CNN layers in parallel, then time-attention mechanism is applied on the CNN outputs, followed by a fully connected layer. The model is trained using unauth chargeback labels. Model structure is depicted in Fig. \ref{fig.supervised_network_structure}.

\begin{figure}[t]
\centering
\includegraphics[height=0.3\textwidth]
{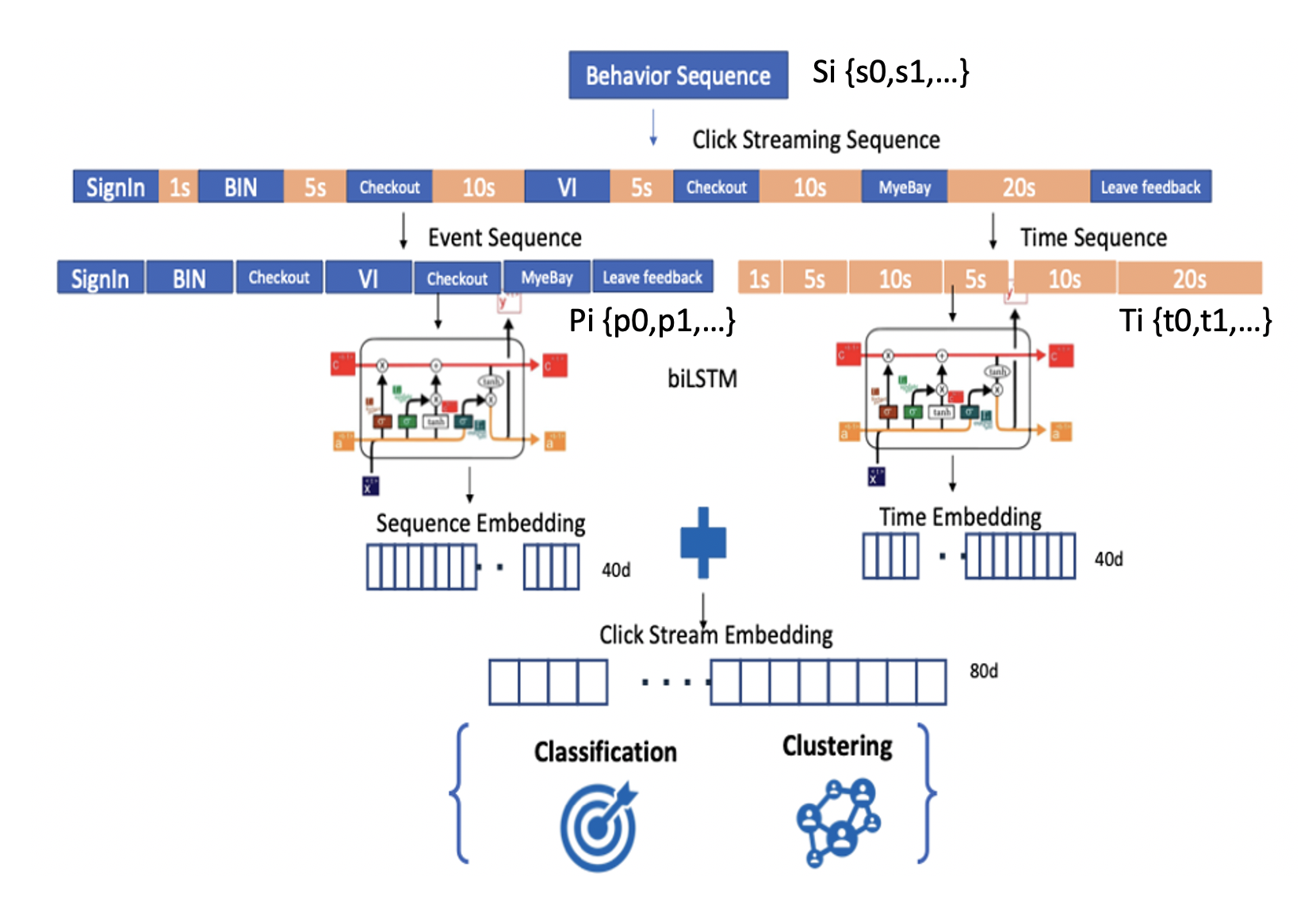}
\caption{Unsupervised time attention LSTM neural network Structure}
\label{fig.unsupervised_LSTM_structure}
\end{figure}

Due to the extremely small fraud population, behavioral embedding trained based on the structure in Fig. \ref{fig.supervised_network_structure} has not significantly improved model performance. 
Thus we try to apply an unsupervised or self-supervised structure named seq2item (shown in Fig. \ref{fig.unsupervised_LSTM_structure}). 
This is the structure we used to generate the sequence embedding for user behavioral similarity measurement.
We prepare the training data inputs as a batch of page view sequences with fixed length of 20 events \{\textbf{\textit{s}}\textsubscript{\textit{i}}\} (\textit{i} = 0, 2, ..., 19). Padding and truncating is applied on short and long sequences. The learning task is designed as predicting the next page ID \{\textbf{\textit{p}}\textsubscript{\textit{20}}\} 
given previous 20 page view information \{\textbf{\textit{s}}\textsubscript{\textit{i}}\}, \textbf{\textit{s}}\textsubscript{\textit{i}} consists of user features such as page ID, page time. As the same page view sequence with different dwell times could be a key factor to differentiate fraud from good users, we add time-attention mechanism in sequence embedding learning. Implementation details can be found in this paper \cite{DBLP:explainableDeepBehaviorSeqClustering}. 

\begin{figure}[t]
\centering
\includegraphics[width=0.4\textwidth]{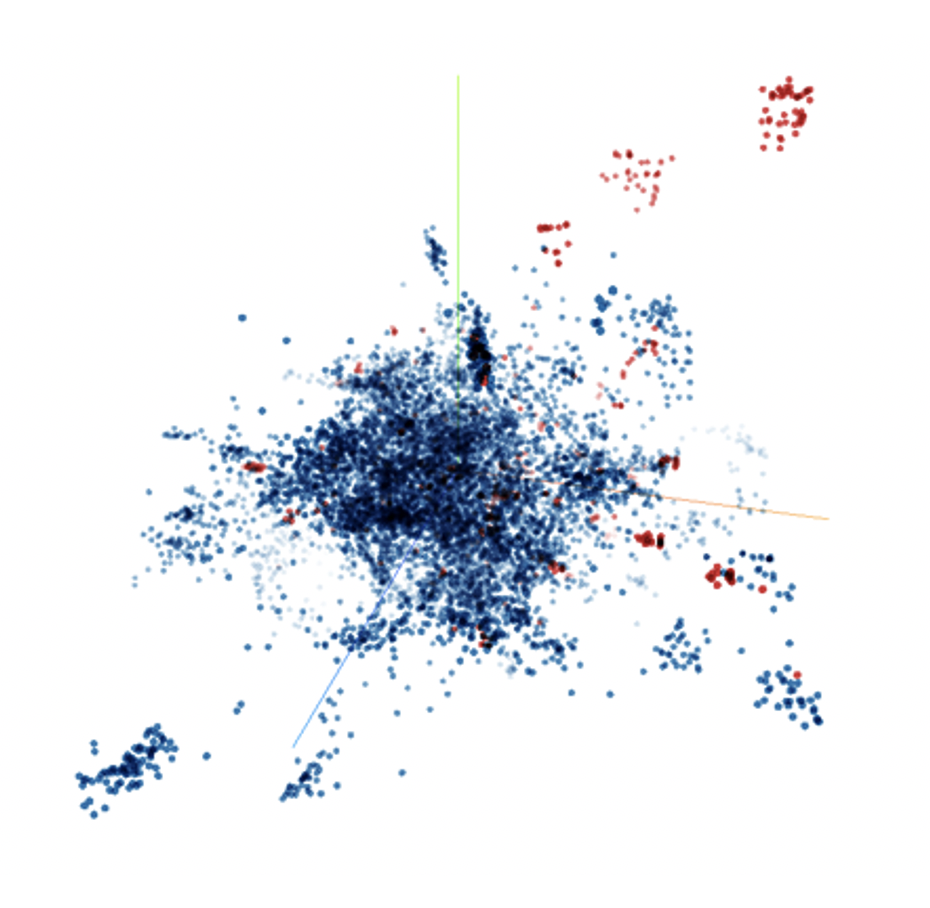}
\caption{Project behavior embedding to 3D space. Red for the fraud transactions, blue ones are the normal transactions. }
\label{fig.tsne}
\end{figure}

Since we don't need labels to train the model, after embeddings are generated on test set, we use two ways to evaluate the embedding efficacy. Firstly we visualize the embeddings using T-SNE. Fig. \ref{fig.tsne} illustrates the result, where red dots are for fraudulent users and blue ones for good users. We can see embedding vectors of fraud users are closer to each other while far away from those of good users.

\begin{figure}[b]
\centering
\includegraphics[width=0.35\textwidth]{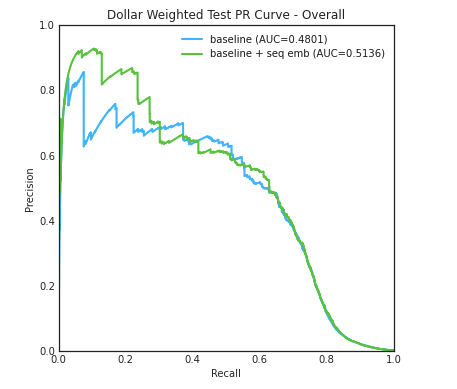}
\caption{Dollar-wise precision recall curve to compare the performance between baseline model vs model trained with baseline and behavior embedding features. The Blue curve is for the baseline model.}
\label{fig.pr_curve}
\end{figure}

On the other hand, we add embedding features to traditional gradient boosted decision trees model as extra behavioral sequence features. From dollar-wise precision-recall perspective, shown in Fig. \ref{fig.pr_curve}, embeddings from unsupervised model do improve the model performance. 

Based on evaluation results above, we choose unsupervised learning for the behavioral sequence embedding part in our graph feature framework.

\subsection{Clustering and build similarity graph}
\begin{figure}[tp]
\centering
\includegraphics[width=0.45\textwidth]{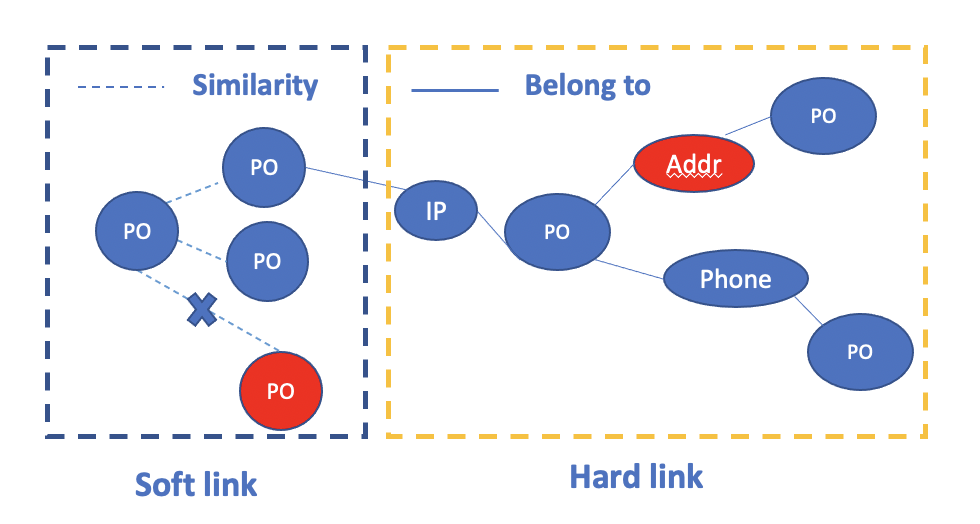}
\caption{Soft link VS Hard link.}
\label{fig.softlink}
\end{figure}

After behavioral sequence embeddings are generated using aforementioned unsupervised method, we can use them to construct the similarity graph. The easiest way is to directly calculate the similarity score between each pair of transactions, then determine whether there should be an edge according to a preset threshold in the experiment.
As shown in Fig. \ref{fig.softlink}, 
different purchase order (PO) could be linked together by their behavior instead of traditional hard linkage link Address(Addr) or Phone, etc. If the similarity exceed preset threshold, then there won't be an edge between this pair of transactions in the graph.
However, the extremely large amount of transactions makes this approach prohibitively expensive, and inevitably introduces a lot of noise. Therefore, we first run a clustering algorithm, and remove isolated nodes and hub nodes based on clustering results. This technique makes the graph building procedure 10x faster than original approach. Here we use in-house GPU based HDBSCAN algorithm for clustering due to its advantages in operational efficiency and performance. In Tab. \ref{tab.cluster_method_compare}, we list the running time and other performance related metrics of different clustering methods.



\begin{table*}[tb]
\centering
    \begin{threeparttable}
    \caption{Different clustering method performance}
    \centering
    \begin{tabular}{cccccc}
    \hline
    Strategy                       & runtime       & \#.Risky C\tnote{a} & Precision & Recall & F-score \\ \hline
    HDBSCAN                        & $\sim$2days   & 31         & 8\%       & 0.20\% & 0.38\%  \\
    PCA(20) $\rightarrow$ HDBSCAN  & 3hours        & 25         & 5\%       & 0.12\% & 0.23\%  \\
    OPTICS                         & 1$\sim$2days  & 45         & 4\%       & 0.17\% & 0.32\%  \\
    Kmeans(k=1200)                 & 5mins per iter\tnote{b} & 25         & 14\%      & 0.58\% & 1.12\%  \\
    pHDBSCAN\tnote{c}              & 5mins         & 37         & 17\%      & 0.39\% & 0.75\%  \\
    \hline
    \end{tabular}
    \label{tab.cluster_method_compare}
    \begin{tablenotes}
        \scriptsize
        \item The comparison is run on the same dataset.
        
        \item[a]\#.Ricky C is the number of risky clusters we have after running above algorithms.
        \item[b]For Kmeans, the runtime depends on iteration, on a average, each iteration takes about 5 mins. More details could be found in  \cite{DBLP:explainableDeepBehaviorSeqClustering}.
        \item[c]pHDBSCAN is our in-house GPU based HDBSCAN application.
    \end{tablenotes}
    \end{threeparttable}
\end{table*}



\begin{figure*}[htbp]
\centering
\includegraphics[width=0.8\textwidth]{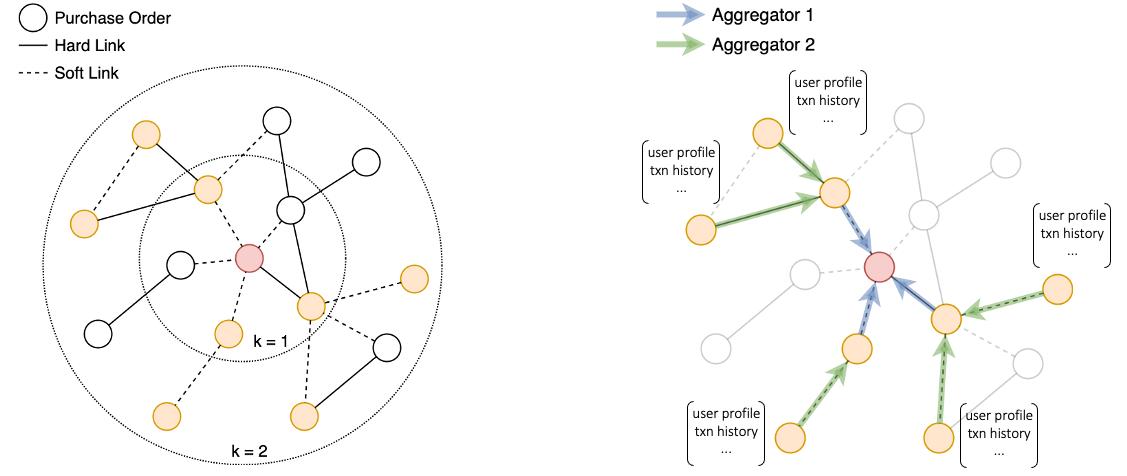}
\caption{GraphSAGE in our application. Each node represents a purchase order, from a more general perspective, we include both soft and hard link in this figure. For feature aggregation part, we generate different domains of feature, such as user profile and transaction history.}
\label{fig.graphSAGE}
\end{figure*}

\subsection{Generate boosting features by unsupervised GraphSage algorithm}
On top of behavioral similarity based graph, we adopt unsupervised GNN algorithm to learn transaction node boosting features, which aggregates features from neighboring transactions and current transaction node itself. Since most of our tasks are inductive, we apply GraphSAGE in our graph embedding framework. The main steps of GraphSAGE in our application have been shown in Fig. \ref{fig.graphSAGE}. At the beginning representations are defined as statistic features of each nodes, such as buyer, seller profiles, transaction histories, denoted as \textbf{h}\rlap{\textsuperscript{0}}\textsubscript{\textit{u}}. At each step \textit{k}, node \textit{v} aggregates the representations of sampled immediate neighboring nodes \{\textbf{h}\rlap{\textsuperscript{k-1}}\textsubscript{\textit{u}}, $\forall$\textit{u}$\in$\textit{N}(\textit{v})\}, then concatenates with the node’s current representation \textbf{h}\rlap{\textsuperscript{k-1}}\textsubscript{\textit{v}}, $\forall$\textit{v}$\in$\textit{V}. This concatenated vector is fed through a fully connected layer with nonlinear activation function $\sigma$, which transforms the representations to be used at the next step. In our method, node embedding features are learned by aggregating neighboring node representations in two steps.
We want to train graph embeddings as a kind of general features applicable to various business scenarios, therefore in this paper, we choose mean aggregator and unsupervised objective. Basically, our target is to minimize the distance to neighbor nodes and increase the distance to non-neighbor nodes. There are two key factors in generating the boosting features. The first one is sampling, based on behavioral similarity graph, we adopt random-walk algorithm to derive a neighborhood sub-graph, then use it as the positive graph, at the same time we randomly sample non-neighbor nodes as negative sub-graph. In real cases, negative samples account for the vast majority, behavior sequences of many negative samples are quite distinctive from positive ones. 
As mentioned in contrastive learning \cite{NEURIPS2020_d89a66c7} \cite{chen2020simple},
selection of negative samples has a great impact on the learning results, so here we choose negative samples based on the similarity or distance of nodes. Another key factor is how many neighbors we want to aggregate. We have tried multiple choices for the number of neighbors above a certain threshold, however the performance did not improve but the time cost increased. So in practice, we use a number a bit less than the threshold.

\subsection{Combine boosting features with existing features}
Existing fraud detection models usually utilize statistic features, such as buyer transaction velocity, user profile, etc. In some specific scenarios, for example, new/guest buyers, lacking historical user information will severely affect the accuracy of model predictions. With boosting features aggregated from neighboring nodes in behavioral graph, we will be able to predict the risk level of these users based on information passed from other users as long as there're some kinds of linkages between them. In this work, we add graph generated embedding feature to traditional tableau features to train a gradient boosting decision tree classification model and evaluate the model performance lift by the graph embedding features.

\section{Experiments}
\subsection{Datasets and Evaluation Metrics}
We use real-world transaction data from a multinational e-commerce website to validate the efficacy of our method. In the experiments, fraudulent transactions are labeled as positive, while good user transactions are negative. The whole dataset is spitted into two separate sets chronologically for training and test.
Training and test dataset both includes about 100 statistics features, for example, buyer and seller's profile features, buyer previous transaction history, etc. More Details of our dataset is summarized in Tab. \ref{tab.dataset}. 

\begin{table}[t]
\caption{Statistics of final train and test datasets}
\centering
\begin{tabular}{cccc}
\hline
Dataset     & \# Positive     & \# Total     & \# Positive rate    \\ \hline
Training    & 54,765 & 1,777,399 & 0.031  \\
Testing & 19,276 & 562,772 & 0.034  \\
 \hline
\end{tabular}
\label{tab.dataset}
\end{table}

Because of the highly imbalanced dataset, we measure model performance by ROC-AUC, PR-AUC and recall at specific precision points. In our data analysis, different user groups show distinctive characteristics and risk levels. Therefore, we carry out extensive experiments with various design choices of graph embedding learning, then evaluate their effects on each user group. More details of user group settings are shown in Tab. \ref{tab.segment}.


\begin{table}[t]
\caption{Transaction/Buyer segments}
\centering
\scriptsize
\begin{tabular}{cccc}
\hline
Group No. & Description                                & \#rate\_in\_total & \#fraud\_rate    \\ \hline
1         & All transactions in test set               & 100.00\%           & 3.20\%         \\
2         & New/guest buyer transactions (cold start)  &   5.89\%           & 16.32\%          \\
3         & Transactions with linkage features         &   30.38\%          & 4.10\%          \\
4         & Ratepay transactions                       &   1.16\%           & 3.82\%         \\ \hline
\end{tabular}
\label{tab.segment}
\end{table}

\subsection{Experimental Results and Analysis} 
As the whole experiment process is complicated, we illustrate our proposed idea into following simplified flow chart in Fig. \ref{fig.pipeline} for better understanding.

\subsubsection{Baseline model}
Our baseline model is a gradient boost tree model trained with 100 statistic features. To make experiment results comparable, our transaction node embedding will be generated based on the same feature set.
\subsubsection{Models with graph embedding features} \hfill 

\begin{table}[t]
\centering
    \begin{threeparttable}
    \caption{Performance of different methods}
    \centering
    \scriptsize
    \begin{tabular}{ccccc p{0.45\textwidth}}
        \hline
        \begin{tabular}[c]{@{}c@{}}Population\\ (AP/AUC)\end{tabular}     & Group 1     & Group 2     & Group 3     & Group 4     \\ \hline
        Baseline                & \textbf{0.361/0.947} & 0.439/0.943          &          0.449/0.958 & 0.579/0.757 \\
        Soft link graph         & 0.341/0.946          & \textbf{0.483}/0.952 &          0.419/0.957 & \textbf{0.589/0.758} \\
        Hard link               & 0.311/0.946          & 0.454/\textbf{0.955} & \textbf{0.786/0.988} & 0.574/0.751 \\ \hline
    \end{tabular}
    \label{tab.performance_method}
    \begin{tablenotes}
        \scriptsize
        \item In each cell, the first metric represents for PR-AUC/average precision and the second one represents for ROC-AUC.
    \end{tablenotes}
    \end{threeparttable}
\end{table}

\begin{table}[t]
\centering
\begin{threeparttable}
\caption{Precision at different recalls and groups}
\centering
\begin{tabular}{lcc p{0.45\textwidth}}
    \hline
              & Group 2 (p@r=0.27) & Group 4 (p@r=0.35) \\ \hline
    Baseline  & 0.82               & 0.80               \\
    Soft link & 0.88               & 0.84               \\ \hline
    \end{tabular}
 \begin{tablenotes}
        \scriptsize
        \item The value in each cell means the precision at this specific recall. It's a common metric to compare different methods in transaction fraud detection scenario.
    \end{tablenotes}
\label{tab.prec_at_diff_rec_and_group}
\end{threeparttable}
\end{table}

\noindent \textbf{Soft link graph} 
   As described in section III, our behavioral graph embedding framework mainly consists of three steps: 
   
   Firstly we use an unsupervised sequence embedding model to generate behavioral embeddings on training and test sets. 
    In this step, we choose the last 20 page views and corresponding dwell times of current checkout session as inputs, use the sequence model described in approach section to generate behavioral embedding for each transaction.

   Then based on behavioral embeddings, we calculate similarities between transactions to establish a soft-link graph. To speed up the graph construction procedure, we leverage GPU-HDBSCAN clustering algorithm to filter out isolated nodes and large clusters. Inspired by GraphSage, we adopt an unsupervised way to learn each transaction node's graph embedding features. As for GraphSage raw features, we choose the same feature set as the baseline model.
In unsupervised GraphSage model, we apply aggregation within 2-hop neighbors, the input feature dimension is 100, consisting of buyer and seller profile, buyer transaction history, etc. As for output embedding dimension, we tried 32 and 64.

%
   Finally, the graph embedding features are used as extra boosting features to the baseline model. According to the model performance results shown in Tab. \ref{tab.performance_method}, for some user groups (ie. Group 2), significant improvement has been achieved after adding boosting features generated from the soft-link graph. Intuitively, new/guest buyers tend to have limited historical data, with the soft-link graph, some of them could be linked with existing users as neighbors so that aggregated features from neighbor transactions help improve the accuracy of risk prediction on these buyers. We also evaluate the precision at certain recall points, which is a widely used metric in 
   e-commerce website online strategies. Here we choose two recall values, firstly at recall of 0.27, we can see from Tab. \ref{tab.prec_at_diff_rec_and_group}, the precision has increased from 0.82 to 0.88. As we try to catch more fraudulent transactions, we increase the recall from 0.27 to 0.35. At that recall threshold, precision of soft-link graph method is also better than the baseline model in Group 2 and Group 5. \\

\noindent \textbf{Hard link} 
   We also test the effectiveness of graph embedding features extracted from a hard-link graph relative to the soft-link graph above. In hard-link graph, connections between transaction nodes are defined by shared properties such as same device, shipping address, etc. Apart from graph construction method, we keep other procedures the same as soft-link graph method. Due to the sparsity of hard-linking strategies, in our dataset, the number of linkages in hard link graph is about 20\% of those in soft link graph. From Tab. \ref{tab.performance_method}, we can see for user Group 2, adding hard-link graph features also results in performance lift on the baseline model although it does not perform better than the soft-link method. But in user Group 3, which only includes transactions linked to each other, performance of hard-link method is much better than the other two methods, this group accounts for less than 3\% of the total population. 

More detailed numbers are displayed in Tab. \ref{tab.performance_method}. \\

\noindent \textbf{Soft and hard link} 
Experiments above show that both hard and soft link embedding features can help improve model performance in specific user groups. In this experiment, we try to combine soft and hard links in graph setup. In addition, since we can control the linkage density of soft link graph through similarity threshold, here we also demonstrate the results from two different similarity thresholds used in soft-link graph. As shown in Tab. \ref{tab.5}, dense refers to smaller threshold, which means we set the similarity score more strict. We can learn that in most groups, loose threshold(larger threshold) cannot achieve better performance than the dense one. This is mainly because when we increase the threshold more transactions would be linked, then more noise might be included. \\

\begin{table}[t]
\begin{threeparttable}
\caption{Performance of different similarity threshold}
\centering
\scriptsize
\begin{tabular}{lcccc p{0.45\textwidth}}
\hline
\begin{tabular}[c]{@{}c@{}}Population (Large)\\ (AP/AUC)\end{tabular} & Group 1     & Group 2     & Group 3     & Group 4     \\ \hline
Soft+hard-loose             & 0.282/0.946 & 0.415/0.955 & 0.295/0.951 & 0.581/\textbf{0.755} \\
Soft+hard-dense             & \textbf{0.306}/0.946 & \textbf{0.436}/0.955 & \textbf{0.638/0.969} & \textbf{0.578}/0.752 \\ \hline
\end{tabular}
\label{tab.5}
\begin{tablenotes}
    \item Different similarity threshold to construct graph for the soft link part, hard linkage remains the same.
\end{tablenotes}
\end{threeparttable}
\end{table}

\noindent \textbf{Different boosting feature dimensions} 
In the graph feature generation step, we also test two different feature dimensions of 32-D and 64-D for soft and hard link graphs. Results in Tab. \ref{tab.6} show that for Group 1 and 2, we don't need a higher dimension, which could speed up the whole pipeline. While for Group 3 and 4, higher dimension performs better. Based on these experiments, we could adjust our dimension choices according to different user populations.\\

\begin{table}[t]
\caption{Performance of different embedding dimension}
\begin{threeparttable}
\centering
\scriptsize
\begin{tabular}{ccccc}
\hline
\begin{tabular}[c]{@{}c@{}}Population (Large)\\ (AP/AUC)\end{tabular} & Group 1     & Group 2     & Group 3     & Group 4     \\ \hline
Boost Feature 32d           & \textbf{0.309}/0.946 & \textbf{0.479}/0.955 & 0.563/0.968 & 0.567/0.750 \\
Boost Feature 64d           & 0.306/0.946 & 0.436/0.955 & \textbf{0.638/0.969} & \textbf{0.578/0.752} \\ \hline
\end{tabular}
\label{tab.6}
\begin{tablenotes}
    \item Experiments with different boosting feature dimensions for soft and hard-link combined graph.
\end{tablenotes}
\end{threeparttable}
\end{table}

\section{Conclusion and future work}
In this work, as far as we know, it's the first time behavior similarity has been used to define graph linkages. Due to the sparsity of hard linking strategies, soft-link graph can help us take advantage of more similar neighbor node features. To speed up the whole process, we leverage clustering method before graph construction. As we generate features by unsupervised GNN, we're able to use these features in more general downstream tasks. Extensive experiments show our approach has achieved better performance compared with baseline model trained with traditional statistic features. 
Especially in new/guest buyer segment  transaction scenario, the precision increased from 0.82 to 0.86 at the same recall of 0.27. 
Having higher precision with the same level of recall enables us to reject less number of good transactions. This way, we could improve our gross merchandise volume and user experience.  


%
In the future, we plan to extend this framework to other risk management scenarios such as user registration, seller payout, etc. In addition, we will try to find a real-time inference solution for this framework so as to detect fraudulent transactions immediately and further improve the user experience. 
\AtNextBibliography{\footnotesize}
\printbibliography{}
\end{document}